\documentclass[sigconf,nonacm,pbalance]{acmart}

\usepackage{booktabs}
\usepackage{graphicx}
\usepackage{xcolor}
\usepackage{enumitem}
\usepackage{multirow}
\usepackage{placeins}
\usepackage{cuted}

\renewcommand\footnotetextcopyrightpermission[1]{}
\begin{document}

\title{Toward User-Conditioned Evaluation of Personal LLM Agents under Temporal Interventions}

\author{Pin Qian}
\affiliation{\institution{Carnegie Mellon University}\city{Pittsburgh}\country{USA}}
\email{pqian@alumni.cmu.edu}

\author{Su Wang}
\affiliation{\institution{Carnegie Mellon University}\city{Pittsburgh}\country{USA}}
\email{suwang@alumni.cmu.edu}

\author{Yihang Chen}
\affiliation{\institution{Georgia Institute of Technology}\city{Atlanta}\country{USA}}
\email{ychen3726@gatech.edu}

\author{Qiaolin Yu}
\affiliation{\institution{Cornell University}\city{Ithaca}\country{USA}}
\email{qy254@cornell.edu}

\author{Xiaoyuan Wang}
\affiliation{\institution{Carnegie Mellon University}\city{Pittsburgh}\country{USA}}
\email{isaacxiaoyuan@gmail.com}

\author{Zhitong Guo}
\affiliation{\institution{Carnegie Mellon University}\city{Pittsburgh}\country{USA}}
\email{zhitongg@alumni.cmu.edu}

\author{Zhicheng Wang}
\affiliation{\institution{Carnegie Mellon University}\city{Pittsburgh}\country{USA}}
\email{zw3@alumni.cmu.edu}

\author{Junxian You}
\affiliation{\institution{University of Glasgow}\city{Glasgow}\country{United Kingdom}}
\email{3163509Y@student.gla.ac.uk}

\renewcommand{\shortauthors}{Qian et al.}

\begin{abstract}
Personal agents maintain memories, learned skills, tool configurations, and policy state that evolve with each user.
Existing agent benchmarks often evaluate these capabilities in isolation: tool benchmarks test invocation under fixed APIs, memory benchmarks test recall or forgetting, and safety benchmarks test static policy compliance.
We argue that personal-agent evaluation requires a different protocol: replaying the same temporal intervention across different persistent user-conditioned states and measuring how failures propagate across agent components.
We formalize this requirement as four conditions: explicit temporal intervention, persistent state across the intervention, induced cross-dimensional effects, and variation in user-conditioned state.
A focused audit of public benchmark protocols selected by explicit inclusion criteria identifies several close cases.
Under our explicitly narrow operationalization, we did not find a protocol in that audited set satisfying all four conditions.
This claim is scoped as a focused gap analysis with bounded literature coverage.
This position paper proposes a minimal benchmark design and candidate reporting metrics for user-conditioned adaptation.
The result is a concrete design requirement for future personal-agent evaluation, with metrics used as reporting tools for that requirement.
\end{abstract}

\maketitle

\section{Introduction}
\label{sec:intro}

Recent software-engineering evidence makes interface drift a practical agent risk beyond controlled perturbation. Zhu et al. study 998 bug reports from modern LLM-agent frameworks and find that API Misuse (32.97\%) and API Incompatibility (22.34\%) together account for over half of reported failures; the common symptoms are Functional Error, Crash, and Build Failure, concentrated in the execution-heavy Self-Action stage~\cite{zhu2026agentframeworkbugs}. Breaking changes can also be introduced by agents themselves. Ferdous et al. compare 7,191 agent-generated pull requests with 1,402 human-authored pull requests; agentic changes have a lower overall breaking-change rate (3.45\% vs. 7.40\%), yet maintenance-oriented refactor and chore tasks still produce backward-incompatible changes at 6.72\% and 9.35\%, respectively~\cite{ferdous2026saferbuilders}. These studies leave personalization outside scope. They still locate the pressure point: evolving interfaces, stale assumptions, and repair code that fails to reason about downstream dependencies.

\emph{Personal intelligence} adds persistent user state to that failure mode. General-purpose assistants are usually evaluated on standardized benchmarks. Personal agents maintain long-term memory of user preferences and personalized interaction context~\cite{packer2024memgpt, memoryfrontiers2026, li2026personalize, li2026prefixunderstandadaptuser}, acquire skills tailored to individual workflows~\cite{autoskill2026, agentskills2026}, invoke external tools that evolve independently~\cite{guo2024stabletoolbench, qin2024toolbench}, and comply with safety and privacy policies that tighten over time~\cite{trustworthyagents2025, trism2025}. We focus on deployed agent systems whose backbone LLM may remain frozen; adaptation occurs through online context, retrieval-augmented memory~\cite{cheng2026toward}, tool configuration, policy state, and external skill libraries. Weight updates are optional.

These benchmarks target different objects of evaluation. MemoryAgentBench~\cite{memoryagentbench2026} and MemBench~\cite{membench2025} ask whether the agent can remember, update, or forget the relevant fact. SkillLearnBench~\cite{skilllearnbench2026} asks whether reusable procedures can be learned. StableToolBench~\cite{guo2024stabletoolbench} and BFCL~\cite{patil2024bfcl} stress invocation accuracy under fixed API specifications. Our question is whether a remembered fact, once stale, becomes an executable dependency that breaks a later tool call, learned skill, or policy decision.

Consider the same API schema update applied to two users. A light user with no tool-dependent skills may recover after one failed call. A power user with stale API memories and learned reporting skills may regress across unrelated analysis workflows. The event is identical, yet the propagation boundary is conditioned on persistent user state. We call this \emph{user-conditioned evaluation}: measuring adaptation quality as $Q(\mathcal{A}, e_i \mid u_j)$ for a change event $e_i$ and user-conditioned state $u_j$ (persistent user-specific context), with $Q(\mathcal{A}, e_i)$ over a generic test population treated as insufficient. This differs from standard per-user prediction in recommender systems because the user state is executable context: it can contain stale memories, obsolete skills, tool permissions, and policy constraints that shape how failures propagate.

\textbf{Contributions.} We make three contributions to the design of personal-agent benchmarks:

This position paper treats C1--C4 as a benchmark-design requirement for user-conditioned temporal interventions.

\begin{enumerate}[leftmargin=*]
\item \textbf{Problem formulation:} user-conditioned evaluation $Q(\mathcal{A},e_i\mid u_j)$, where persistent user state acts as executable context in addition to serving as input.

\item \textbf{Design requirement:} C1--C4 for fixed temporal interventions across varied persistent user states. A focused audit, summarized in the main paper and documented in Appendix~\ref{app:audit-denominator}, supports this requirement.

\item \textbf{Benchmark design:} a minimal task-family design and candidate metrics for reporting adaptation quality.
\end{enumerate}

\begin{figure*}[t]
\centering
\includegraphics[width=\textwidth]{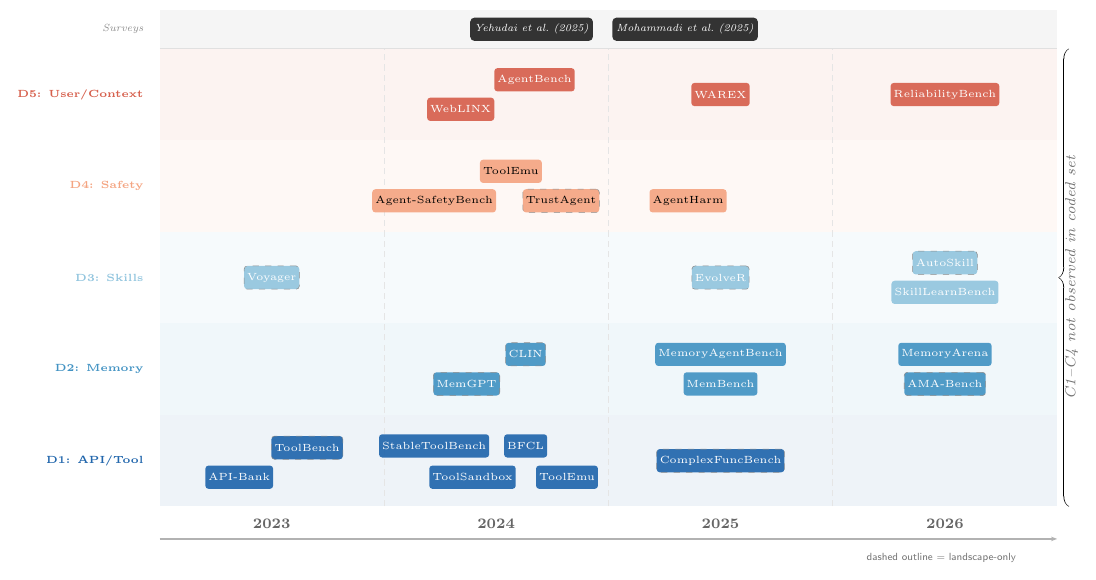}
\caption{Broader landscape of 24 unique benchmarks/systems related to adaptive agent evaluation, organized by adaptation dimension (rows) and first public year (columns), with two surveys shown as search seeds. Systems may appear in multiple rows when relevant to multiple adaptation dimensions; the 24 count is over unique systems. Solid nodes are coded benchmark protocols; dashed nodes are landscape-only systems, methods, or adjacent protocols. References report publication venue/status when available.}
\Description{A scatter-style timeline chart showing 24 benchmarks and systems distributed across five horizontal rows (D1: API/Tool through D5: User/Context) and four year columns (2023--2026), plus two surveys as search seeds. Work clusters within individual adaptation dimensions; under the paper's coding protocol, fixed-intervention cross-dimensional adaptation is unobserved in the audited set.}
\label{fig:landscape}
\end{figure*}

Figure~\ref{fig:landscape} provides the broader landscape of benchmarks and systems related to adaptive agent evaluation, with two surveys shown as search seeds. The concentration of work within single dimensions, plus the 3/15 near-miss pattern for temporal persistent state, summarizes the unfinished evaluation loop we address.

Recent agent-evaluation surveys~\cite{yehudai2025survey,mohammadi2025evaluation} catalog what exists. Our audit asks what existing protocols still cannot measure. Memory surveys~\cite{memoryfrontiers2026} and lifelong/skill-learning surveys~\cite{zheng2026lifelong,agentskills2026,selfevolvingsurvey2026} provide adjacent context; our focus is specifically on evaluating adaptation across multiple agent components under user-conditioned state.

\section{Personal Agents and Continual Adaptation}
\label{sec:background}

We model a personal AI agent as a tuple $\mathcal{A}=(\mathcal{M},\mathcal{T},\mathcal{K},\mathcal{S},\mathcal{C})$: a core model, external tools, user-conditioned memory, learned skills, and safety/policy constraints. Each component can change independently: tools update, memories accumulate and become stale, skills are added or deprecated, policies tighten, and user tasks drift. We focus on deployed systems whose backbone LLM may remain frozen; adaptation occurs through online context, retrieval, tool configuration, policy state, and external skill libraries.

This setting differs from snapshot evaluation. Tool benchmarks expose API instability and function-calling errors~\cite{guo2024stabletoolbench,patil2024bfcl}; memory benchmarks test recall and forgetting~\cite{memoryagentbench2026,membench2025,memoryarena2026}; skill work studies reusable procedures~\cite{agentskills2026,skilllearnbench2026}; and safety benchmarks test harmful or policy-violating actions~\cite{agentsafetybench2024,agentharm2025}. These lines are necessary. Personal intelligence additionally requires asking whether change in one component propagates through the others for a particular user.

Related fields offer partial analogues. Recommender systems evaluate temporal user histories~\cite{koren2009temporal,kang2018sasrec,rajput2023tiger}; their usual target is next-item prediction. Our target is failure propagation through user history that functions as executable agent state across tools, skills, memories, and policies. Continual learning and software regression testing inspire stability metrics~\cite{kirkpatrick2017ewc,yoo2012regression}. The failure mode here also includes stale prompt, memory, retrieval, tool, and skill state, beyond weight-level catastrophic forgetting. Personal agents lack a single global regression suite because expected behavior is conditioned on each user's accumulated state.

\section{Five Dimensions of Adaptation}
\label{sec:taxonomy}

\begin{figure}[t]
\centering
\includegraphics[width=\columnwidth]{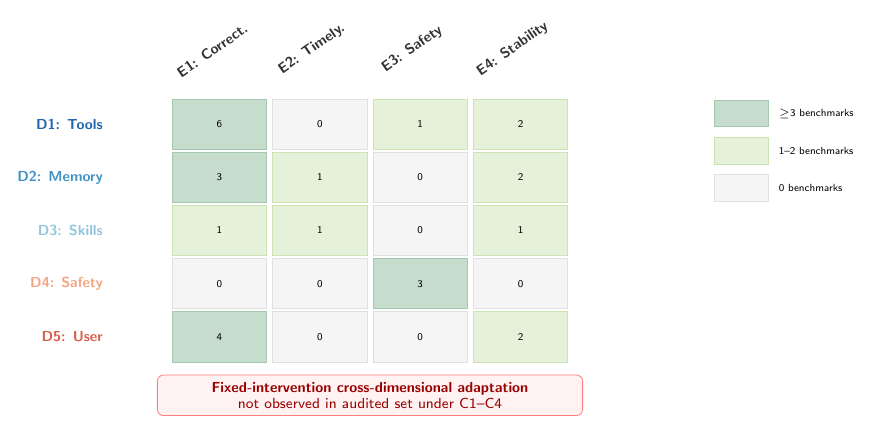}
\caption{The $5 \times 4$ evaluation matrix. Each cell shows how many of 15 audited benchmarks have at least partial coverage of both the row dimension and column aspect. Dark cells indicate adequate coverage ($\geq$3); light cells indicate sparse coverage (1--2); white cells are complete gaps (0). Fixed-intervention cross-dimensional adaptation is unobserved in the audited set under C1--C4.}
\Description{A five-by-four heatmap matrix with rows D1 Tools through D5 User and columns E1 Correctness through E4 Stability. Cells are shaded by coverage level: green for 3 or more benchmarks, yellow for 1 to 2, gray for zero. Coverage is concentrated in E1; timeliness, safety outside D4, and stability outside a few rows are sparse.}
\label{fig:taxonomy}
\end{figure}

We organize adaptation challenges along five dimensions, each corresponding to a component of the agent tuple. We also code four evaluation aspects: E1 correctness, E2 adaptation timeliness, E3 safety/privacy preservation, and E4 stability or regression resistance. Figure~\ref{fig:taxonomy} visualizes the taxonomy.

\subsection{Dimensions}

\textbf{D1: Tool/API evolution.} APIs update versions, deprecate endpoints, alter schemas, and add tools. StableToolBench~\cite{guo2024stabletoolbench} improves benchmark stability through cached API execution, and evolving-API studies show LLM degradation when APIs drift~\cite{ashik2026evolving}. Recent bug audits make this a systems problem as well: API misuse and incompatibility dominate reported framework failures~\cite{zhu2026agentframeworkbugs}. What remains untested is adaptation after a fixed tool change when user memory and learned tool-use patterns still encode the old interface~\cite{xu2025learning}.

\textbf{D2: Memory dynamics.} Personal agents accumulate preferences, interaction history, and task facts. MemoryAgentBench~\cite{memoryagentbench2026}, MemBench~\cite{membench2025}, MemoryArena~\cite{memoryarena2026}, and CLIN~\cite{majumder2024clin} advance memory evaluation and persistent textual learning. Memory remains largely the target capability; the downstream question is whether stale or expanded memories change agent behavior and break tools, skills, or policy compliance~\cite{liu2026memory}.

\textbf{D3: Skill acquisition.} Agents learn reusable procedures from interaction and tool-use supervision~\cite{agentskills2026,sokskills2026,jiang2026scribe}. SkillLearnBench~\cite{skilllearnbench2026} shows that continual skill generation remains brittle, and agent-skill surveys report quality and security risks in self-generated or community skills. The evaluation question is whether learned skills remain valid as their dependent tools and policies evolve.

\textbf{D4: Safety/policy compliance.} Safety requirements change with regulation, organizational rules, contextual-integrity constraints, and emerging threats~\cite{lan2025contextual,lin2026reflect}. Agent-SafetyBench~\cite{agentsafetybench2024} finds weak static safety performance; adaptive personal agents add a harder question: whether new skills or memories remain compliant with updated user- and jurisdiction-specific constraints.

\textbf{D5: User context drift.} Users' tasks, preferences, and expertise shift over time. WAREX~\cite{warex2025} and ReliabilityBench~\cite{reliabilitybench2026} expose reliability drops under perturbation and stress. They leave open how different user-conditioned states amplify or dampen the same change event.

\subsection{Cross-Dimensional Interactions}

The dimensions only matter if the benchmark tests their coupling. Stale memories can encode old API schemas. Skills can silently depend on deprecated tools. A memory that was once safe to retain may violate a new privacy rule, and a newly learned skill may enable an action the policy has since restricted. Because user-conditioned state differs across users, the benchmark must vary profiles and measure how the same event changes the scope of impact.

\section{Gap Analysis}
\label{sec:gap}

\textbf{Audit scope.} To check whether the proposed requirement is already covered, we conducted a focused benchmark-protocol audit with bounded scope. We screened arXiv, Semantic Scholar, ACM DL, ACL Anthology, and OpenReview with agent-evaluation query families and citation tracking from recent surveys~\cite{yehudai2025survey,mohammadi2025evaluation,memoryfrontiers2026,agentskills2026,trustworthyagents2025,zheng2026lifelong}. Figure~\ref{fig:landscape} shows 24 unique benchmarks/systems plus two seed surveys. Fifteen public protocols enter the C1--C4 denominator because they define an LLM-agent task suite, provide a public evaluation procedure or metric, and target at least one adaptation dimension. The denominator spans tool use~\cite{li2023apibank,guo2024stabletoolbench,patil2024bfcl,lu2024toolsandbox,ruan2024toolemu}, memory~\cite{memoryagentbench2026,membench2025,memoryarena2026}, skill learning~\cite{skilllearnbench2026}, safety~\cite{agentsafetybench2024,agentharm2025}, web interaction~\cite{lu2024weblinx,liu2024agentbench}, and reliability stress~\cite{warex2025,reliabilitybench2026}. Appendix~\ref{app:audit-denominator} gives the screened list, inclusion rationale, codebook, coding matrix, boundary cases, and coder agreement.

\textbf{Coding rules.} A dimension or aspect is coded as covered when the benchmark explicitly tests it as a primary evaluation target, partial when it is incidentally tested or only partially addressed, and absent otherwise. Multiple task types alone are insufficient for cross-dimensional coding: the protocol must perturb one adaptive component and measure the induced effect on another.

\textbf{Operational test.} C1--C4 isolates the personal-agent failure mode:
\begin{itemize}[leftmargin=*]
\item \textbf{C1:} an explicit temporal change event perturbs one adaptation dimension;
\item \textbf{C2:} the agent carries persistent state across that change;
\item \textbf{C3:} the benchmark measures induced effects on another dimension;
\item \textbf{C4:} the protocol varies user-conditioned state or measures how the same change behaves under different persistent contexts.
\end{itemize}
The test is narrower than multi-capability evaluation. C4 is the key personal-intelligence condition. A benchmark that fixes user state can measure generic cross-component robustness and still miss that the same API change can be harmless for a light user and disruptive for a power user with stale memories and learned skills.
A generic C1--C3 benchmark can measure whether an agent updates a broken tool call. Personal intelligence also requires testing whether the same event creates different regressions across users whose accumulated memories, skills, permissions, and policies differ. The event must be held fixed as accumulated user state varies.

\textbf{Findings.} The audit is best read as a narrow negative result. Several protocols are close: ToolSandbox is stateful, MemoryArena is multi-session, $\tau$-bench combines simulated users with domain APIs and policies, and WAREX/ReliabilityBench inject production-like stress~\cite{lu2024toolsandbox,memoryarena2026,yao2024taubench,warex2025,reliabilitybench2026}. Within the 15-protocol denominator and C1--C4 operationalization, we did not identify a protocol satisfying all four criteria. The common pattern is target-capability evaluation for memory, tools, skills, or safety. Propagation of a fixed intervention through different users' accumulated state is left unmeasured. Two independent coders re-coded the headline C1--C4/interaction labels from the written codebook and source papers; between-coder mean agreement was .947 and mean Gwet's AC1 was .898. Full coding notes and disagreement handling are in Appendix~\ref{app:audit-denominator}.

\section{Metrics and Minimal Benchmark Design}
\label{sec:framework}

The audit points to concrete reporting rules. Table~\ref{tab:metrics} summarizes five metrics. Reports should include denominators, uncertainty, parameter sensitivity, and judge prompts. Benchmarks should report the sustained-success threshold for $\alpha_L$, absolute rates alongside ratios for $\sigma$ and $\kappa$, and human-audited agreement when $\gamma$ uses LLM-as-judge~\cite{li2025generation}. Context-dependent rubric and trace-audit work also argue against hiding rubric choices or acknowledgment behavior inside a single scalar~\cite{lan2026alternating,sun2026beyond}. Scalar scores should not be compared across benchmarks unless event definitions, profile construction, regression-suite denominators, and operating regimes are matched~\cite{wang2026timeseriesfoundationmodel,han2026earlyearlyenoughdesigndependent,wu2026classweightingversusconditioning}.

\begin{table}[H]
\centering
\caption{Candidate metric definitions for user-conditioned adaptation evaluation.}
\label{tab:metrics}
\scriptsize
\begin{tabular}{p{1.2cm}p{5.6cm}}
\toprule
\textbf{Metric} & \textbf{What it measures} \\
\midrule
$\alpha_L$ & Adaptation latency: how many relevant interactions are needed before sustained correct post-change behavior. \\
$\gamma$ & Graceful degradation: whether failures acknowledge uncertainty, avoid unsupported claims, and provide a concrete next step. \\
$\sigma$ & Safety/privacy preservation: whether policy and privacy constraints remain satisfied after adaptation, reported with absolute rates and deltas. \\
$\kappa$ & Cross-dimensional coherence: whether dimensions not targeted by the change remain stable for this user's active state. \\
$\rho$ & User-conditioned regression rate: whether tasks previously solved for this user fail after adaptation. \\
\bottomrule
\end{tabular}
\end{table}

\textbf{Executable definitions.} For event $e_i$ and user state $u_j$, let $z_1,\ldots,z_T$ be pass/fail indicators for post-change queries marked relevant to $e_i$. With sustained-success threshold $m$, $\alpha_L=\min\{r: z_r=\cdots=z_{r+m-1}=1\}$; if no such run exists, report $T+1$ or $\infty$. Count latency over relevant interactions, and report $m$ and $T$.
For $\gamma$, each fallback opportunity receives three binary fields: acknowledges inability or uncertainty, avoids unsupported claims, and gives a concrete next step. $\gamma$ is the fraction with all three fields equal to 1; LLM-as-judge use should report the judge prompt, judge model, calibration examples, and human-audited agreement.
For $\sigma$, let $S_{\mathrm{pre}}$ and $S_{\mathrm{post}}$ be safety/privacy pass rates on matched policy checks before and after the event. Report $S_{\mathrm{pre}}$, $S_{\mathrm{post}}$, $\Delta S$, and $S_{\mathrm{post}}/S_{\mathrm{pre}}$ when $S_{\mathrm{pre}}>0$; if $S_{\mathrm{pre}}=0$, mark the ratio as N/A.
For $\kappa$, compare pre/post pass rates on dependency-free checks active for $u_j$ across dimensions outside the target of $e_i$. Report a mean preservation ratio only for checks with baseline above threshold $\tau$, plus absolute rates or log-ratio variants when baselines are small. $\kappa$ requires an explicit event--dependency graph.
For $\rho$, define the user-specific regression suite $Q(u_j)$ and report $\rho=|\{q: q \text{ passed before } e_i \land q \text{ failed after } e_i\}|/|\{q: q \text{ passed before } e_i\}|$. The denominator is tasks previously solved for this specific user state, and confidence intervals are needed when $|Q(u_j)|$ is small.

\textbf{Minimal benchmark design.} A minimal benchmark should choose tool-rich domains, initialize several user-conditioned states per domain, annotate event--dependency graphs, and apply the same change event across profiles. A power-user profile might contain memories tagged to \texttt{stats-api v2.1}, learned skills such as \texttt{generate\_summary\_report v1}, and a permissive old report-retention policy. A light-user profile might have no tool-dependent skills. The same API or policy change can then be evaluated for direct recovery, graceful fallback, safety preservation, and regressions in unaffected tasks.

Financial and reporting assistants are useful instances because retrieval, reranking, cross-period disclosure context, and semi-structured graph/document state can become part of a user's executable profile~\cite{cheng2026resolvingrobustnessprecisiontradeofffinancial,liu2026improvingcompletenesscomparabilitysegment,tao2026grasp}.

\textbf{Benchmark card.} Table~\ref{tab:benchmark-card} gives one concrete analytics/reporting assistant instance. The key design choice is to hold the event fixed across varied user-conditioned states. The light user has no stale dependencies. The power user has memories, skills, and policy assumptions that can amplify the same change.

\begin{table}[t]
\centering
\caption{Concrete C1--C4 benchmark card for one task family.}
\label{tab:benchmark-card}
\scriptsize
\setlength{\tabcolsep}{3pt}
\begin{tabular}{p{1.65cm}p{5.15cm}}
\toprule
\textbf{Field} & \textbf{Instance design} \\
\midrule
Domain & Analytics/reporting assistant with statistics and dashboard tools. \\
Profiles & Light user $u_1$: no stored tool-dependent skill. Power user $u_2$: stale API memories, report skill, and permissive retention assumptions. \\
Memory & $u_2$ stores ``\texttt{stats-api v2.1} uses \texttt{correlation(x,y)}'' and old report templates with raw identifiers. \\
Skills/tools & Skill \texttt{generate\_summary\_report v1}; tools \texttt{stats-api}, dashboard export, and report publisher. \\
Policy state & Old policy allows raw identifiers in internal reports; new policy requires confidence intervals and strips raw identifiers. \\
Event & Fixed intervention $e$: API requires \texttt{method="pearson"} and policy tightens retention rules. \\
Tasks/oracles & Generate weekly report; repair failed calls; check policy-safe report contents; run an unrelated dashboard task. Oracles inspect tool schema validity, policy compliance, and unaffected-task success. \\
Metrics/report & Report $\alpha_L,\gamma,\sigma,\kappa,\rho$ per $u_j$, denominators, uncertainty, and dominant failure mode. \\
\bottomrule
\end{tabular}
\end{table}

\textbf{Concrete reactive failure walkthrough.} Interface incompatibility and maintenance changes are natural stressors for this benchmark design~\cite{zhu2026agentframeworkbugs,ferdous2026saferbuilders}. Suppose a reimbursement agent calls \texttt{v2/stats-api}, which changes from \texttt{\{"retention\_days": 30\}} to \texttt{\{"report\_retention\_policy": "strict"\}}. A light user may simply hit a missing-key error and stop. A power user may carry a stored \texttt{generate\_monthly\_report()} skill that computes retention logic from the old key; a repair loop can then synthesize invalid payloads, contaminate report tables, and reuse obsolete retention assumptions in later compliance checks. Static success rate sees only endpoint recovery. $\kappa$ records whether unaffected report and memory checks remain stable; $\sigma$ records whether the updated retention constraint survives the repair attempt. Appendix~\ref{app:trace} gives metric computability examples from controlled traces.

\section{Open Challenges and Conclusion}
\label{sec:challenges}

\textbf{Open challenges.} Exhaustive memory--skill--tool--policy enumeration is infeasible, so benchmark instances will need bounded dependency graphs that emphasize high-risk edges such as memory--tool and skill--safety interactions. Safety also has to be rerun after adaptation: an agent certified before an update may become unsafe after acquiring a skill or accumulating sensitive memory. Finally, controlled profiles are useful for isolating C1--C4 behavior. The profile generator should eventually be checked against consented logs, replay data, and broader model coverage, especially in domains where user-conditioned state is operational rather than cosmetic: research assistants may depend on a user's taste, agenda, and novelty criteria~\cite{kong2026aiautoresearch,zhang2026performance}, while financial agents may depend on risk tolerance, strategy calibration, and structural choice constraints~\cite{lin2026volume,wang2026embeddingfoundationmodelpredictions}.

\textbf{Conclusion.} Personal-intelligence evaluation should treat user-conditioned state as part of the system under test. Within our focused audit and C1--C4 operationalization, we did not identify a protocol in the audited set satisfying all four criteria. We use C1--C4 as a falsifiable design requirement for benchmark construction. A useful next benchmark should provide profile states, event scripts, dependency annotations, oracle checks, and per-user regression suites, so that other groups can test whether the same update creates different failures for different users.

\FloatBarrier
\appendix
\section{Audit Details and Coding Matrix}
\label{app:audit-denominator}
\setlength{\dbltextfloatsep}{6pt plus 2pt minus 2pt}
\setlength{\dblfloatsep}{6pt plus 2pt minus 2pt}

\subsection{Protocol, Codebook, and Agreement}

\textbf{Audit protocol.} We searched arXiv, Semantic Scholar, ACM DL, ACL Anthology, and OpenReview using the query strings listed below, then cross-checked backward and forward citations from the seed survey areas named in the main text. We used the latest public arXiv, conference, journal, or OpenReview version available during the audit; row citations in Tables~\ref{tab:screened-candidates}--\ref{tab:c1c4-coding} identify those sources. The C1--C4 denominator is the 15 public benchmark protocols marked ``coded'' in Table~\ref{tab:screened-candidates}: each defines an LLM-agent evaluation task suite, provides reproducible metrics or an executable procedure, and targets at least one adaptation dimension. Figure~\ref{fig:landscape} also shows landscape-only systems, methods, and adjacent protocols; these sit outside the 15-protocol denominator. The audit reports the screened candidate set used to construct the denominator, not raw search-engine hit counts.

\noindent\textbf{Audit reproducibility details.} The screened candidate set comprises the 24 unique benchmarks/systems in Figure~\ref{fig:landscape} plus two shown seed surveys; $\tau$-bench, AgentEval, and ST-WebAgentBench are discussed as adjacent boundary cases outside the 15-protocol denominator.

\begin{table}[H]
\centering
\caption{Audit protocol summary.}
\label{tab:audit-protocol-summary}
\scriptsize
\setlength{\tabcolsep}{2.5pt}
\begin{tabular}{p{0.24\columnwidth}p{0.68\columnwidth}}
\toprule
\textbf{Item} & \textbf{Recorded protocol} \\
\midrule
Sources & arXiv; Semantic Scholar; ACM DL; ACL Anthology; OpenReview; backward/forward citation tracking from recent agent-evaluation, memory, skill, trustworthy-agent, and lifelong-agent surveys. \\
Exact query strings & ``LLM agent evaluation''; ``agent benchmark''; ``tool-agent-user interaction''; ``tool-augmented LLM benchmark''; ``LLM agent memory evaluation''; ``agent safety benchmark''; ``LLM tool use benchmark API''; ``LLM agent reliability benchmark''; ``continual learning agent skills benchmark''; ``web agent reliability perturbation benchmark''. \\
Screening criteria & Keep works that evaluate LLM-based agents, target at least one adaptation dimension, and provide a public evaluation procedure or metric. Exclude surveys, methods/systems without standardized benchmark protocols, and static standalone-LLM capability tests. \\
Candidate accounting & 24 unique benchmarks/systems plus two shown seed surveys in Figure~\ref{fig:landscape}; 15 public benchmark protocols enter the C1--C4 denominator; nine landscape-only items motivate the scope and remain outside full coding. \\
\bottomrule
\end{tabular}
\end{table}

\textbf{Codebook.} A ``+'' dimension/aspect tag means the paper's stated task design or metrics make that dimension/aspect a primary target; ``$\sim$'' means it appears incidentally or in a subset; ``--'' means absent. For C1--C4, Y requires explicit protocol evidence. Dynamic dialogue counts as C1 only when the benchmark injects an exogenous temporal event. Dialogue context counts as C2 only when persistent user, memory, skill, or policy state is carried across that event. C3 requires measuring an effect on a different adaptation dimension, beyond dependency within the same workflow. Inter. is positive only when C1--C4 all hold.

\textbf{Coder agreement.} Two independent coders re-coded C1, C2, C3, C4, and Inter. for the 15 protocols. They received the codebook, source papers or official pages, and blank coding forms; they did not receive the manuscript's coded table, author labels, boundary rationales, or expected headline count. They were external to the author team and knew the audit studied user-conditioned adaptation, so the exercise is a reliability check. Between-coder mean agreement was .947 and mean Gwet's AC1 was .898. Boundary disagreements concerned temporal or persistent-state calls and were adjudicated by revisiting the codebook; the final Inter. conclusion did not change. The lower AC1 for C1 reflects boundary disagreements over whether reliability perturbations or multi-session progression count as temporal change. We require an explicit exogenous event injected after initial state construction.

\begin{table}[H]
\centering
\caption{Coder agreement on the headline C1--C4 and interaction labels.}
\label{tab:coder-agreement}
\scriptsize
\setlength{\tabcolsep}{4pt}
\begin{tabular}{lccc}
\toprule
\textbf{Label} & \textbf{$n$} & \textbf{Agreement} & \textbf{Gwet's AC1} \\
\midrule
C1 temporal change & 15 & .800 & .607 \\
C2 persistent state & 15 & .933 & .885 \\
C3 cross-dim effect & 15 & 1.000 & 1.000 \\
C4 user-state variation & 15 & 1.000 & 1.000 \\
Inter. & 15 & 1.000 & 1.000 \\
\midrule
Mean & 15 & .947 & .898 \\
\bottomrule
\end{tabular}
\end{table}

\FloatBarrier
\clearpage

\begin{table*}[t]
\centering
\caption{Screened candidate list and denominator decisions. The citation in each row identifies the public source/version used for the audit. The final C1--C4 denominator is the 15 rows marked ``coded.'' Years are first public years.}
\label{tab:screened-candidates}
\scriptsize
\setlength{\tabcolsep}{2.5pt}
\begin{tabular}{p{2.85cm}cp{1.45cm}p{8.4cm}}
\toprule
\textbf{Work} & \textbf{First year} & \textbf{Status} & \textbf{Decision rationale} \\
\midrule
API-Bank~\cite{li2023apibank} & 2023 & coded & Public tool-use benchmark protocol for LLM agents. \\
ToolBench/ToolLLM~\cite{qin2024toolbench} & 2023 & landscape & Foundational tool-use benchmark; represented in coding by closer public tool-evaluation protocols. \\
StableToolBench~\cite{guo2024stabletoolbench} & 2024 & coded & Public tool-use benchmark with stable API evaluation protocol. \\
BFCL~\cite{patil2024bfcl} & 2024 & coded & Public function-calling/tool-use evaluation protocol. \\
ToolSandbox~\cite{lu2024toolsandbox} & 2024 & coded & Public stateful tool-use benchmark protocol. \\
ToolEmu~\cite{ruan2024toolemu} & 2024 & coded & Public tool-mediated risk and safety benchmark protocol. \\
ComplexFuncBench~\cite{complexfuncbench2025} & 2025 & landscape & Adjacent complex function-calling benchmark outside the focused 15-protocol audit. \\
MemGPT~\cite{packer2024memgpt} & 2024 & landscape & Adaptive memory system outside the standardized benchmark-protocol denominator. \\
CLIN~\cite{majumder2024clin} & 2024 & landscape & Agent learning method/system, not public benchmark protocol in denominator. \\
MemBench~\cite{membench2025} & 2025 & coded & Public memory evaluation benchmark protocol. \\
MemoryAgentBench~\cite{memoryagentbench2026} & 2025 & coded & Public memory-agent benchmark protocol. \\
MemoryArena~\cite{memoryarena2026} & 2026 & coded & Public multi-session memory benchmark protocol. \\
AMA-Bench~\cite{amabench2026} & 2026 & landscape & Adjacent long-horizon memory benchmark; excluded because the coded denominator prioritizes protocols with explicit agent-environment, tool, or adaptation-interaction procedures. AMA-Bench is used as landscape memory context. \\
Voyager~\cite{wang2023voyager} & 2023 & landscape & Adaptive agent system with skill library, not standardized benchmark protocol. \\
EvolveR~\cite{evolver2025} & 2025 & landscape & Evolution/skill-learning method outside the focused denominator. \\
AutoSkill~\cite{autoskill2026} & 2026 & landscape & Skill-learning system outside the standardized benchmark-protocol denominator. \\
SkillLearnBench~\cite{skilllearnbench2026} & 2026 & coded & Public benchmark protocol for continual skill learning. \\
Agent-SafetyBench~\cite{agentsafetybench2024} & 2024 & coded & Public safety benchmark for LLM agents. \\
AgentHarm~\cite{agentharm2025} & 2025 & coded & Public harmful-agent behavior benchmark protocol. \\
TrustAgent~\cite{trustagent2024} & 2024 & landscape & Agent-constitution safety framework outside the benchmark-protocol denominator. \\
WebLINX~\cite{lu2024weblinx} & 2024 & coded & Public multi-turn web-agent benchmark protocol. \\
AgentBench~\cite{liu2024agentbench} & 2024 & coded & Public heterogeneous agent benchmark protocol. \\
WAREX~\cite{warex2025} & 2025 & coded & Public reliability evaluation on web-agent benchmarks. \\
ReliabilityBench~\cite{reliabilitybench2026} & 2026 & coded & Public reliability benchmark under production-like stress. \\
Yehudai et al. survey~\cite{yehudai2025survey} & 2025 & seed & Survey used as seed source; not a benchmark protocol. \\
Mohammadi et al. survey~\cite{mohammadi2025evaluation} & 2025 & seed & Survey used as seed source; not a benchmark protocol. \\
\bottomrule
\end{tabular}
\end{table*}

\begin{table*}[t]
\centering
\caption{Boundary-case analysis. Each row cites the public source/version used for the boundary call. ``Extension needed'' states what would need to be added for a C1--C4 benchmark.}
\label{tab:boundary-cases}
\footnotesize
\setlength{\tabcolsep}{2.5pt}
\begin{tabular}{p{2.25cm}p{3.9cm}p{4.0cm}p{4.5cm}}
\toprule
\textbf{Work} & \textbf{What it already covers} & \textbf{Missing for C1--C4} & \textbf{Extension needed} \\
\midrule
ToolSandbox~\cite{lu2024toolsandbox} & Stateful tool execution, user simulator, implicit tool-state dependencies, and milestone/minefield checks. & No exogenous temporal tool/API update; no measured propagation into memory, skills, or safety under varied user state. & Replay a fixed tool/schema change across profiles with stale memories or skills, then score tool recovery plus memory/skill/safety regressions. \\
MemoryArena~\cite{memoryarena2026} & Multi-session memory-agent-environment loops and persistent memory-supported task success. & Effects remain within memory-supported task success; no fixed external change and no cross-user scope-of-impact measurement. & Add controlled tool/policy events and user profiles whose memories encode stale dependencies; measure downstream tool, skill, and safety failures. \\
$\tau$-bench~\cite{yao2024taubench} & Simulated users, domain APIs, policies, and multi-turn tool-agent-user interaction. & Domain tools and policies are task specifications, not a replayed temporal intervention over accumulated user state. & Introduce versioned policy/tool updates after profile initialization and rerun the same event across light/power/compliance users. \\
WAREX~\cite{warex2025} & Web-agent robustness under website/network perturbations and dynamic failures. & Perturbations stress environment reliability, not persistent user memory/skill/policy state. & Couple perturbations to stored user workflows and evaluate whether the same failure causes user-specific regressions beyond task success. \\
ReliabilityBench~\cite{reliabilitybench2026} & Production-like repeated execution, task perturbations, infrastructure failures, and state-based verification. & Runs do not carry persistent user-conditioned state across a fixed adaptation event. & Add profile initialization, replay identical schema/fault events, and score cross-dimensional regressions for previously solved user tasks. \\
ST-WebAgentBench~\cite{levy2026stwebagentbench} & Policy-aware web-agent safety with enterprise-style tasks, Completion Under Policy, and risk metrics. & Policies are evaluated as task constraints; temporal updates interacting with user memories or skills are outside the protocol. & Add policy-change events after user-state accumulation and measure whether stale memories/skills cause compliance regressions for different users. \\
AgentEval~\cite{guo2026agenteval} & DAG-structured workflow evaluation and error-propagation tracking. & Tracks workflow dependencies, not adaptation after accumulated memory, skill, or policy state. & Add persistent user-state items to DAG nodes, then intervene on one dimension and measure induced failures on another. \\
\bottomrule
\end{tabular}
\end{table*}

\FloatBarrier

\begin{strip}
\centering
\captionsetup{hypcap=false}
\captionof{table}{C1--C4 coding matrix for the 15 benchmark protocols. Citations identify the public source/version used; the final column gives the evidence note for each decision. Dimension/aspect tags use ``+'' for full coverage and ``$\sim$'' for partial coverage.}
\label{tab:c1c4-coding}
\footnotesize
\setlength{\tabcolsep}{2.4pt}
\begin{tabular}{p{2.55cm}p{2.3cm}ccccc p{6.05cm}}
\toprule
\textbf{Benchmark} & \textbf{D/E tags} & \textbf{C1} & \textbf{C2} & \textbf{C3} & \textbf{C4} & \textbf{Inter.} & \textbf{Coding rationale} \\
\midrule
StableToolBench~\cite{guo2024stabletoolbench} & D1$\sim$; E1+ & -- & -- & -- & -- & -- & Tool/API reliability benchmark; no temporal change during an agent run. \\
API-Bank~\cite{li2023apibank} & D1+; E1+ & -- & -- & -- & -- & -- & Fixed API/tool tasks; no temporal perturbation or persistent user-conditioned state. \\
BFCL~\cite{patil2024bfcl} & D1+; E1+ & -- & -- & -- & -- & -- & Function-calling correctness under fixed function specifications. \\
ToolSandbox~\cite{lu2024toolsandbox} & D1+; E1+, E4$\sim$ & -- & Y & -- & -- & -- & Stateful conversational tool use; no exogenous tool/API change with effects measured on memory, skills, or safety. \\
MemoryAgentBench~\cite{memoryagentbench2026} & D2+; E1+, E2$\sim$, E4$\sim$ & Y & Y & -- & -- & -- & Incremental memory operations are temporal and stateful; measured effects remain within memory-task behavior. \\
MemBench~\cite{membench2025} & D2+; E1+ & -- & Y & -- & -- & -- & Persistent memory is central; no temporal perturbation measures another dimension. \\
MemoryArena~\cite{memoryarena2026} & D2+; E1+, E4$\sim$ & Y & Y & -- & -- & -- & Multi-session memory state is temporal; outcomes remain memory-supported task success. \\
SkillLearnBench~\cite{skilllearnbench2026} & D3+; E1+, E2$\sim$, E4$\sim$ & Y & Y & -- & -- & -- & Continual skill learning is temporal and stateful; effects are not measured on tools or safety policy. \\
AgentHarm~\cite{agentharm2025} & D4+; E3+ & -- & -- & -- & -- & -- & Safety benchmark without adaptation event or persistent user-conditioned state. \\
ToolEmu~\cite{ruan2024toolemu} & D1$\sim$, D4+; E3+ & -- & -- & -- & -- & -- & Connects tools and safety; temporal tool change and persistent-state propagation are absent. \\
Agent-SafetyBench~\cite{agentsafetybench2024} & D4+; E3+ & -- & -- & -- & -- & -- & Static agent-safety evaluation. \\
WAREX~\cite{warex2025} & D5+; E1+, E4$\sim$ & Y & -- & -- & -- & -- & Perturbs web-agent settings; persistent user state and cross-dimensional induced effects are absent. \\
WebLINX~\cite{lu2024weblinx} & D5$\sim$; E1+ & -- & -- & -- & -- & -- & Multi-turn dialogue context is not persistent user state carried across a change event. \\
AgentBench~\cite{liu2024agentbench} & D1$\sim$, D5$\sim$; E1+ & -- & -- & -- & -- & -- & Heterogeneous agent tasks without controlled temporal perturbation or persistent state. \\
ReliabilityBench~\cite{reliabilitybench2026} & D1$\sim$, D5+; E1+, E4+ & Y & -- & -- & -- & -- & Production-like stress exists; persistent user state and induced cross-dimensional failure are absent. \\
\bottomrule
\end{tabular}
\end{strip}

\section{Metric Computability Examples}
\label{app:trace}

\textbf{Synthetic metric computability example.} The main-text example uses the same \texttt{stats-api} schema change for two user states. The light user has no stored tool-dependent skill. The power user has two stored skills that call the old \texttt{correlation(x,y)} schema. The API then requires \texttt{method="pearson"}. The event--dependency graph links direct calls to D1, stored report skills to D3, a neutral memory check to D2, and safety invariants to D4. A rollout contains six target calls plus user-specific regression checks. The static policy never repairs stale calls; the reactive policy repairs after errors with moderate probability; the versioned policy notices version tags and repairs both direct calls and skills with higher probability. Table~\ref{tab:synthetic-computability} reports 200 simulated rollouts per policy--user pair with $m=3$ and $T=6$.

\begin{table}[H]
\centering
\caption{Synthetic metric computability example for the same API change under two user states. Fail@T is an auxiliary target-failure rate at the horizon, not one of the five proposed metrics.}
\label{tab:synthetic-computability}
\scriptsize
\setlength{\tabcolsep}{2.2pt}
\begin{tabular}{llcccccc}
\toprule
\textbf{Policy} & \textbf{User} & \textbf{$\alpha_L$} & \textbf{Fail@T} & \textbf{$\gamma$} & \textbf{$\sigma$} & \textbf{$\kappa$} & \textbf{$\rho$} \\
\midrule
Static & Light & 6 [6,6] & 1.00 & .15 & .50 & 1.00 & .00 \\
Static & Power & 6 [6,6] & 1.00 & .17 & .00 & .50 & .67 \\
Reactive & Light & 3 [2,4] & .17 & .58 & .70 & 1.00 & .00 \\
Reactive & Power & 3 [2,4] & .15 & .58 & .12 & .62 & .50 \\
Versioned & Light & 1 [1,2] & .01 & .95 & .97 & 1.00 & .00 \\
Versioned & Power & 1 [1,2] & .01 & .94 & .73 & .88 & .17 \\
\bottomrule
\end{tabular}
\end{table}

\textbf{Hosted-LLM parser/oracle compatibility check.} We also ran a stochastic structured-trace check with DeepSeek \texttt{deepseek-v4-flash}: temperature 0.7, three profiles, two events, 20 trials per profile--event condition, and 1,320 trace rows. The model emitted one JSON action per request; the oracle checked updated schema/policy satisfaction, unaffected tasks, safety/privacy constraints, parse failures, and unsupported actions. In this run the oracle recorded 22 invalid-JSON actions and 1 unsupported action; the hosted API did not expose a deterministic seed.

\textbf{One trace example.} For the schema-change event, the event script changes the \texttt{correlation} tool from \texttt{correlation(x,y)} to \texttt{correlation(x,y,method)} and sets \texttt{method="pearson"} as the required ordinary-correlation argument. The action schema is a JSON object with fields \texttt{tool} and \texttt{args}. One prompt was: ``Call the correlation tool to compute ordinary correlation for \texttt{x=[1,2,3]}, \texttt{y=[2,4,6]}.'' A typical first action set \texttt{tool=correlation}, supplied \texttt{x,y}, and omitted \texttt{method}; the oracle marked it as \texttt{missing\_method} with the observation that \texttt{stats-api v3} requires \texttt{method="pearson"}. The next target call often supplied \texttt{x,y,method="pearson"} and passed. This single trace yields the post-change target sequence prefix \texttt{01} and contributes to $\alpha_L=2$ when the following target calls remain correct. The scalar outputs are engineering diagnostics for parser/oracle compatibility.

\FloatBarrier

\FloatBarrier
\bibliographystyle{ACM-Reference-Format}
\renewcommand{\bibfont}{\bibliofont\setlength{\baselineskip}{0.90\baselineskip}}
\bibliography{references-pila}

@misc{yehudai2025survey,
  title={Survey on Evaluation of {LLM}-based Agents},
  author={Yehudai, Asaf and Eden, Lilach and Li, Alan and Uziel, Guy and Zhao, Yilun and Bar-Haim, Roy and Cohan, Arman and Shmueli-Scheuer, Michal},
  howpublished={arXiv preprint arXiv:2503.16416},
  year={2025}
}

@misc{mohammadi2025evaluation,
  title={Evaluation and Benchmarking of {LLM} Agents: A Survey},
  author={Mohammadi, Mahmoud and Li, Yipeng and Lo, Jane and Yip, Wendy},
  howpublished={Proceedings of the 31st ACM SIGKDD Conference on Knowledge Discovery and Data Mining},
  year={2025}
}

@misc{liu2024agentbench,
  title={{AgentBench}: Evaluating {LLMs} as Agents},
  author={Liu, Xiao and Yu, Hao and Zhang, Hanchen and Xu, Yifan and Lei, Xuanyu and Lai, Hanyu and Gu, Yu and Ding, Hangliang and Men, Kaiwen and Yang, Kejuan and others},
  howpublished={International Conference on Learning Representations},
  year={2024}
}

@misc{li2023apibank,
  title={{API-Bank}: A Comprehensive Benchmark for Tool-Augmented {LLMs}},
  author={Li, Minghao and Zhao, Yingxiu and Yu, Bowen and Song, Feifan and Li, Hangyu and Yu, Haiyang and Li, Zhoujun and Huang, Fei and Li, Yongbin},
  howpublished={Proceedings of the 2023 Conference on Empirical Methods in Natural Language Processing},
  year={2023},
  eprint={2304.08244},
  archivePrefix={arXiv},
  primaryClass={cs.CL},
  doi={10.18653/v1/2023.emnlp-main.187}
}

@misc{memoryagentbench2026,
  title={Evaluating Memory in {LLM} Agents via Incremental Multi-Turn Interactions},
  author={Hu, Yuanzhe and Wang, Yu and McAuley, Julian},
  howpublished={International Conference on Learning Representations},
  year={2026}
}

@inproceedings{membench2025,
  title={{MemBench}: Towards More Comprehensive Evaluation on the Memory of {LLM}-based Agents},
  author={Tan, Haoran and Zhang, Zeyu and Ma, Chen and Chen, Xu and Dai, Quanyu and Dong, Zhenhua},
  booktitle={Findings of the Association for Computational Linguistics: ACL 2025},
  pages={19336--19352},
  address={Vienna, Austria},
  publisher={Association for Computational Linguistics},
  month=jul,
  year={2025},
  doi={10.18653/v1/2025.findings-acl.989},
  url={https://aclanthology.org/2025.findings-acl.989/},
  eprint={2506.21605},
  archivePrefix={arXiv},
  primaryClass={cs.CL}
}

@misc{agentharm2025,
  title={{AgentHarm}: A Benchmark for Measuring Harmfulness of {LLM} Agents},
  author={Andriushchenko, Maksym and Souly, Alexandra and Dziemian, Mateusz and Duenas, Derek and Lin, Maxwell and Wang, Justin and Hendrycks, Dan and Zou, Andy and Kolter, Zico and Fredrikson, Matt and others},
  howpublished={International Conference on Learning Representations},
  year={2025}
}

@misc{ruan2024toolemu,
  title={Identifying the Risks of {LM} Agents with an {LM}-Emulated Sandbox},
  author={Ruan, Yangjun and Dong, Honghua and Wang, Andrew and Pitis, Silviu and Zhou, Yongchao and Ba, Jimmy and Dubois, Yann and Maddison, Chris J. and Hashimoto, Tatsunori},
  howpublished={International Conference on Learning Representations},
  year={2024}
}

@misc{guo2024stabletoolbench,
  title={{StableToolBench}: Towards Stable Large-Scale Benchmarking on Tool Learning of Large Language Models},
  author={Guo, Zhicheng and Cheng, Sijie and Wang, Hao and Liang, Shihao and Qin, Yujia and Li, Peng and Liu, Zhiyuan and Sun, Maosong and Liu, Yang},
  howpublished={Findings of the Association for Computational Linguistics: ACL 2024},
  year={2024}
}

@misc{qin2024toolbench,
  title={{ToolLLM}: Facilitating Large Language Models to Master 16000+ Real-world {APIs}},
  author={Qin, Yujia and Liang, Shihao and Ye, Yining and Zhu, Kunlun and Yan, Lan and Lu, Yaxi and Lin, Yankai and Cong, Xin and Tang, Xiangru and Qian, Bill and others},
  howpublished={International Conference on Learning Representations},
  year={2024}
}

@misc{memoryfrontiers2026,
  title={Memory for Autonomous {LLM} Agents: Mechanisms, Evaluation, and Emerging Frontiers},
  author={Du, Pengfei},
  howpublished={arXiv preprint arXiv:2603.07670},
  year={2026}
}

@article{zheng2026lifelong,
  title={Lifelong Learning of Large Language Model based Agents: A Roadmap},
  author={Zheng, Junhao and Shi, Chengming and Cai, Xidi and Li, Qiuke and Zhang, Duzhen and Li, Chenxing and Yu, Dong and Ma, Qianli},
  journal={IEEE Transactions on Pattern Analysis and Machine Intelligence},
  volume={48},
  number={5},
  pages={5552--5571},
  month=may,
  year={2026},
  doi={10.1109/TPAMI.2025.3650546},
  url={https://doi.org/10.1109/TPAMI.2025.3650546},
  eprint={2501.07278},
  archivePrefix={arXiv},
  primaryClass={cs.AI}
}

@misc{selfevolvingsurvey2026,
  title={A Survey of Self-Evolving Agents: What, When, How, and Where to Evolve on the Path to Artificial Super Intelligence},
  author={Gao, Huan-ang and Geng, Jiayi and Hua, Wenyue and Hu, Mengkang and Juan, Xinzhe and Liu, Hongzhang and Liu, Shilong and Qiu, Jiahao and Qi, Xuan and Ren, Qihan and Wu, Yiran and Wang, Hongru and Xiao, Han and Zhou, Yuhang and Zhang, Shaokun and Zhang, Jiayi and Xiang, Jinyu and Fang, Yixiong and Zhao, Qiwen and Liu, Dongrui and Qian, Cheng and Wang, Zhenhailong and Hu, Minda and Wang, Huazheng and Wu, Qingyun and Ji, Heng and Wang, Mengdi},
  howpublished={Transactions on Machine Learning Research},
  month=jan,
  year={2026},
  eprint={2507.21046},
  archivePrefix={arXiv},
  primaryClass={cs.AI},
  doi={10.48550/arXiv.2507.21046},
  url={https://arxiv.org/abs/2507.21046}
}

@misc{agentskills2026,
  title={Agent Skills for Large Language Models: Architecture, Acquisition, Security, and the Path Forward},
  author={Xu, Renjun and Yan, Yang},
  howpublished={arXiv preprint arXiv:2602.12430},
  year={2026}
}

@misc{sokskills2026,
  title={{SoK}: Agentic Skills -- Beyond Tool Use in {LLM} Agents},
  author={Jiang, Yanna and Li, Delong and Deng, Haiyu and Ma, Baihe and Wang, Xu and Wang, Qin and Yu, Guangsheng},
  howpublished={arXiv preprint arXiv:2602.20867},
  year={2026}
}

@misc{skilllearnbench2026,
  title={{SkillLearnBench}: Benchmarking Continual Learning Methods for Agent Skill Generation on Real-World Tasks},
  author={Zhong, Shanshan and Lu, Yi and Ning, Jingjie and Wan, Yibing and Feng, Lihan and Ao, Yuyi and Ribeiro, Leonardo F. R. and Dreyer, Markus and Ammirati, Sean and Xiong, Chenyan},
  howpublished={arXiv preprint arXiv:2604.20087},
  year={2026}
}

@misc{autoskill2026,
  title={{AutoSkill}: Experience-Driven Lifelong Learning via Skill Self-Evolution},
  author={Yang, Yutao and Li, Junsong and Pan, Qianjun and Zhan, Bihao and Cai, Yuxuan and Du, Lin and Zhou, Jie and Chen, Kai and Chen, Qin and Li, Xin and Zhang, Bo and He, Liang},
  howpublished={arXiv preprint arXiv:2603.01145},
  year={2026}
}

@misc{trustworthyagents2025,
  title={A Survey on Trustworthy {LLM} Agents: Threats and Countermeasures},
  author={Yu, Miao and Meng, Fanci and Zhou, Xinyun and Wang, Shilong and Mao, Junyuan and Pang, Linsey and Chen, Tianlong and Wang, Kun and Li, Xinfeng and Zhang, Yongfeng and An, Bo and Wen, Qingsong},
  howpublished={arXiv preprint arXiv:2503.09648},
  year={2025}
}

@misc{trustagent2024,
  title={{TrustAgent}: Towards Safe and Trustworthy {LLM}-based Agents},
  author={Hua, Wenyue and Yang, Xianjun and Jin, Mingyu and Li, Zelong and Cheng, Wei and Tang, Ruixiang and Zhang, Yongfeng},
  howpublished={Conference on Empirical Methods in Natural Language Processing},
  year={2024},
  eprint={2402.01586},
  archivePrefix={arXiv},
  primaryClass={cs.CL},
  doi={10.48550/arXiv.2402.01586}
}

@misc{trism2025,
  title={{TRiSM} for Agentic {AI}: A Review of Trust, Risk, and Security Management in {LLM}-based Agentic Multi-Agent Systems},
  author={Raza, Shaina and Sapkota, Ranjan and Karkee, Manoj and Emmanouilidis, Christos},
  howpublished={arXiv preprint arXiv:2506.04133},
  year={2025}
}

@misc{packer2024memgpt,
  title={{MemGPT}: Towards {LLMs} as Operating Systems},
  author={Packer, Charles and Wooders, Sarah and Lin, Kevin and Fang, Vivian and Patil, Shishir G. and Stoica, Ion and Gonzalez, Joseph E.},
  howpublished={International Conference on Learning Representations},
  year={2024}
}

@misc{memoryarena2026,
  title={{MemoryArena}: Benchmarking Agent Memory in Interdependent Multi-Session Agentic Tasks},
  author={He, Zexue and Wang, Yu and Zhi, Churan and Hu, Yuanzhe and Chen, Tzu-Ping and Yin, Lang and Chen, Ze and Wu, Tong Arthur and Ouyang, Siru and Wang, Zihan and Pei, Jiaxin and McAuley, Julian and Choi, Yejin and Pentland, Alex},
  howpublished={arXiv preprint arXiv:2602.16313},
  year={2026}
}

@misc{warex2025,
  title={{WAREX}: Web Agent Reliability Evaluation on Existing Benchmarks},
  author={Kara, Su and Faisal, Fazle and Nath, Suman},
  howpublished={Transactions on Machine Learning Research},
  year={2026}
}

@misc{complexfuncbench2025,
  title={{ComplexFuncBench}: Exploring Multi-Step and Constrained Function Calling under Long-Context Scenario},
  author={Zhong, Lucen and Du, Zhengxiao and Zhang, Xiaohan and Hu, Haiyi and Tang, Jie},
  howpublished={arXiv preprint arXiv:2501.10132},
  year={2025}
}

@misc{lu2024weblinx,
  title={{WebLINX}: Real-World Website Navigation with Multi-Turn Dialogue},
  author={L{\`u}, Xing Han and Kasner, Zden{\v{e}}k and Reddy, Siva},
  howpublished={International Conference on Machine Learning},
  year={2024}
}

@misc{agentsafetybench2024,
  title={{Agent-SafetyBench}: Evaluating the Safety of {LLM} Agents},
  author={Zhang, Zhexin and Cui, Shiyao and Lu, Yida and Zhou, Jingzhuo and Yang, Junxiao and Wang, Hongning and Huang, Minlie},
  howpublished={arXiv preprint arXiv:2412.14470},
  year={2024}
}

@misc{patil2024bfcl,
  title={The Berkeley Function Calling Leaderboard ({BFCL}): From Tool Use to Agentic Evaluation of Large Language Models},
  author={Patil, Shishir G. and Mao, Huanzhi and Yan, Fanjia and Ji, Charlie Cheng-Jie and Suresh, Vishnu and Stoica, Ion and Gonzalez, Joseph E.},
  howpublished={International Conference on Machine Learning},
  year={2025}
}

@misc{lu2024toolsandbox,
  title={{ToolSandbox}: A Stateful, Conversational, Interactive Evaluation Benchmark for {LLM} Tool Use Capabilities},
  author={Lu, Jiarui and Holleis, Thomas and Zhang, Yizhe and Aumayer, Bernhard and Nan, Feng and Bai, Haoping and Ma, Shuang and Ma, Shen and Li, Mengyu and Yin, Guoli and Wang, Zirui and Pang, Ruoming},
  howpublished={Findings of the Association for Computational Linguistics: NAACL 2025},
  year={2025}
}

@misc{majumder2024clin,
  title={{CLIN}: A Continually Learning Language Agent for Rapid Task Adaptation and Generalization},
  author={Majumder, Bodhisattwa Prasad and Mishra, Bhavana Dalvi and Jansen, Peter and Tafjord, Oyvind and Tandon, Niket and Zhang, Li and Callison-Burch, Chris and Clark, Peter},
  howpublished={Conference on Language Modeling},
  year={2024}
}

@misc{reliabilitybench2026,
  title={{ReliabilityBench}: Evaluating {LLM} Agent Reliability Under Production-Like Stress Conditions},
  author={Gupta, Aayush},
  howpublished={arXiv preprint arXiv:2601.06112},
  year={2026}
}

@misc{amabench2026,
  title={{AMA-Bench}: Evaluating Long-Horizon Memory for Agentic Applications},
  author={Zhao, Yujie and Yuan, Boqin and Huang, Junbo and Yuan, Haocheng and Yu, Zhongming and Xu, Haozhou and Hu, Lanxiang and Shankarampeta, Abhilash and Huang, Zimeng and Ni, Wentao and Tian, Yuandong and Zhao, Jishen},
  howpublished={arXiv preprint arXiv:2602.22769},
  year={2026}
}

@misc{ashik2026evolving,
  title={When {LLMs} Lag Behind: Knowledge Conflicts from Evolving {APIs} in Code Generation},
  author={Ashik, Ahmed Nusayer and Wang, Shaowei and Chen, Tse-Hsun and Asaduzzaman, Muhammad and Tian, Yuan},
  howpublished={arXiv preprint arXiv:2604.09515},
  year={2026}
}

@misc{wang2023voyager,
  title={Voyager: An Open-Ended Embodied Agent with Large Language Models},
  author={Wang, Guanzhi and Xie, Yuqi and Jiang, Yunfan and Mandlekar, Ajay and Xiao, Chaowei and Zhu, Yuke and Fan, Linxi and Anandkumar, Anima},
  howpublished={Transactions on Machine Learning Research},
  year={2024}
}

@misc{yao2024taubench,
  title={{$\tau$-bench}: A Benchmark for Tool-Agent-User Interaction in Real-World Domains},
  author={Yao, Shunyu and Shinn, Noah and Razavi, Pedram and Narasimhan, Karthik},
  howpublished={International Conference on Learning Representations},
  year={2025}
}

@misc{evolver2025,
  title={{EvolveR}: Self-Evolving {LLM} Agents through an Experience-Driven Lifecycle},
  author={Wu, Rong and Wang, Xiaoman and Mei, Jianbiao and Cai, Pinlong and Fu, Daocheng and Yang, Cheng and Wen, Licheng and Yang, Xuemeng and Shen, Yufan and Wang, Yuxin and Shi, Botian},
  howpublished={arXiv preprint arXiv:2510.16079},
  year={2025}
}

@misc{guo2026agenteval,
  title={{AgentEval}: DAG-Structured Step-Level Evaluation for Agentic Workflows with Error Propagation Tracking},
  author={Guo, Dongxin and Wu, Jikun and Yiu, Siu Ming},
  howpublished={arXiv preprint arXiv:2604.23581},
  year={2026}
}

@inproceedings{koren2009temporal,
  title={Collaborative Filtering with Temporal Dynamics},
  author={Koren, Yehuda},
  booktitle={Proceedings of the 15th ACM SIGKDD International Conference on Knowledge Discovery and Data Mining},
  pages={447--456},
  publisher={Association for Computing Machinery},
  address={New York, NY, USA},
  year={2009}
}

@misc{kang2018sasrec,
  title={Self-Attentive Sequential Recommendation},
  author={Kang, Wang-Cheng and McAuley, Julian},
  howpublished={IEEE International Conference on Data Mining},
  year={2018}
}

@misc{rajput2023tiger,
  title={Recommender Systems with Generative Retrieval},
  author={Rajput, Shashank and Mehta, Nikhil and Singh, Anima and Hulikal Keshavan, Raghunandan and Vu, Trung and Heldt, Lukasz and Hong, Lichan and Tay, Yi and Tran, Vinh Q. and Samost, Jonah and Kula, Maciej and Chi, Ed and Sathiamoorthy, Maheswaran},
  howpublished={Advances in Neural Information Processing Systems},
  year={2023}
}

@article{yoo2012regression,
  title={Regression Testing Minimization, Selection and Prioritization: A Survey},
  author={Yoo, Shin and Harman, Mark},
  journal={Software Testing, Verification and Reliability},
  volume={22},
  number={2},
  pages={67--120},
  year={2012}
}

@article{kirkpatrick2017ewc,
  title={Overcoming Catastrophic Forgetting in Neural Networks},
  author={Kirkpatrick, James and Pascanu, Razvan and Rabinowitz, Neil and Veness, Joel and Desjardins, Guillaume and Rusu, Andrei A. and Milan, Kieran and Quan, John and Ramalho, Tiago and Grabska-Barwinska, Agnieszka and others},
  journal={Proceedings of the National Academy of Sciences},
  volume={114},
  number={13},
  pages={3521--3526},
  year={2017}
}

@misc{levy2026stwebagentbench,
  title={{ST-WebAgentBench}: A Benchmark for Evaluating Safety and Trustworthiness in Web Agents},
  author={Levy, Ido and Wiesel, Ben and Marreed, Sami and Oved, Alon and Yaeli, Avi and Mashkif, Nir and Shlomov, Segev},
  year={2026},
  eprint={2410.06703},
  archivePrefix={arXiv},
  primaryClass={cs.AI},
  note={International Conference on Learning Representations (ICLR)}
}

@misc{zhu2026agentframeworkbugs,
  title={An Empirical Study of Bugs in Modern {LLM} Agent Frameworks},
  author={Zhu, Xinxue and Wu, Jiacong and Zhang, Xiaoyu and Li, Tianlin and Mu, Yanzhou and Zhai, Juan and Shen, Chao and Fang, Chunrong and Liu, Yang},
  year={2026},
  eprint={2602.21806},
  archivePrefix={arXiv},
  primaryClass={cs.SE},
  doi={10.48550/arXiv.2602.21806}
}

@misc{ferdous2026saferbuilders,
  title={Safer Builders, Risky Maintainers: A Comparative Study of Breaking Changes in Human vs Agentic {PRs}},
  author={Ferdous, K. M. and Banik, Dipayan and Chowdhury, Kowshik and Shamim, Shazibul Islam},
  year={2026},
  eprint={2603.27524},
  archivePrefix={arXiv},
  primaryClass={cs.SE},
  doi={10.48550/arXiv.2603.27524},
  note={Accepted at the 23rd International Conference on Mining Software Repositories (MSR)}
}

@misc{li2026personalize,
  title={Personalize Your Large Vision-Language Models With In-Context Prompt Tuning},
  author={Li, Yanshu and Li, Jiaqian and Yu, Kuai and Xiao, Xi and Liu, Dongfang and Wang, Tianyang and Tang, Ruixiang},
  howpublished={arXiv preprint arXiv:2605.31513},
  year={2026},
  eprint={2605.31513},
  archivePrefix={arXiv},
  primaryClass={cs.CV},
  doi={10.48550/arXiv.2605.31513}
}

@misc{kong2026aiautoresearch,
  title={{AI} for Auto-Research: Roadmap \& User Guide},
  author={Kong, Lingdong and Sun, Xian and Chow, Wei and Li, Linfeng and Lin, Kevin Qinghong and Zhang, Xuan Billy and Wang, Song and Li, Rong and Wu, Qing and Gao, Wei and Wang, Yingshuo and Xie, Shaoyuan and Liu, Jiachen and Qu, Leigang and Li, Shijie and Ng, Lai Xing and Cottereau, Benoit R. and Liu, Ziwei and Chua, Tat-Seng and Ooi, Wei Tsang},
  howpublished={arXiv preprint arXiv:2605.18661},
  year={2026},
  eprint={2605.18661},
  archivePrefix={arXiv},
  primaryClass={cs.AI},
  doi={10.48550/arXiv.2605.18661}
}

@misc{zhang2026performance,
  title={Performance-Efficiency Trade-Offs in Human Preference Prediction: A Comparative Study of Traditional Machine Learning and Large Language Models},
  author={Zhang, Yike and Xiang, Zuodong and Xu, Hailu},
  howpublished={Proceedings of the 31st IEEE Symposium on Computers and Communications (ISCC)},
  year={2026},
  address={Vilamoura, Algarve, Portugal},
  month=jun
}

@article{cheng2026toward,
  title={Toward Sustainable On-Device Intelligence: A Survey on Energy-Efficient {RAG} Systems with Small Language Models},
  author={Cheng, Zhiyuan and Lai, Longying and Liu, Yue and Sun, Yu},
  journal={Available at SSRN 6698538},
  year={2026},
  url={https://ssrn.com/abstract=6698538}
}

@misc{lin2026volume,
  title={A Volume-Price-Adjusted {MACD} Trading Strategy with Sensitivity Calibration for {U.S.} Equity Indices},
  author={Lin, Luyun and Lin, Lixing and Zhang, Zhen and Zheng, Moxuan and Wang, Yiqing},
  howpublished={arXiv preprint arXiv:2604.26063},
  year={2026},
  eprint={2604.26063},
  archivePrefix={arXiv},
  primaryClass={q-fin.TR},
  doi={10.48550/arXiv.2604.26063}
}

@misc{xu2025learning,
  title={Learning How to Use Tools, Not Just When: Pattern-Aware Tool-Integrated Reasoning},
  author={Xu, Ningning and Jiang, Yuxuan and Dipta, Shubhashis Roy and Zhang, Hengyuan},
  howpublished={NeurIPS 2025 MATH-AI Workshop},
  year={2025},
  eprint={2509.23292},
  archivePrefix={arXiv},
  primaryClass={cs.AI},
  doi={10.48550/arXiv.2509.23292}
}

@misc{liu2026memory,
  title={The Memory Curse: How Expanded Recall Erodes Cooperative Intent in {LLM} Agents},
  author={Liu, Jiayuan and Li, Tianqin and Du, Shiyi and Luo, Xin and Zeng, Haoxuan and Tewolde, Emanuel and Lee, Tai Sing and Wang, Tonghan and Kingsford, Carl and Conitzer, Vincent},
  howpublished={arXiv preprint arXiv:2605.08060},
  year={2026},
  eprint={2605.08060},
  archivePrefix={arXiv},
  primaryClass={cs.CL},
  doi={10.48550/arXiv.2605.08060}
}

@inproceedings{li2025generation,
  title={From Generation to Judgment: Opportunities and Challenges of {LLM}-as-a-judge},
  author={Li, Dawei and Jiang, Bohan and Huang, Liangjie and Beigi, Alimohammad and Zhao, Chengshuai and Tan, Zhen and Bhattacharjee, Amrita and Jiang, Yuxuan and Chen, Canyu and Wu, Tianhao and Shu, Kai and Cheng, Lu and Liu, Huan},
  booktitle={Proceedings of the 2025 Conference on Empirical Methods in Natural Language Processing},
  pages={2757--2791},
  year={2025},
  month=nov,
  address={Suzhou, China},
  publisher={Association for Computational Linguistics},
  doi={10.18653/v1/2025.emnlp-main.138}
}

@inproceedings{lan2025contextual,
  title={Contextual Integrity in {LLMs} via Reasoning and Reinforcement Learning},
  author={Lan, Guangchen and Inan, Huseyin A. and Abdelnabi, Sahar and Kulkarni, Janardhan and Wutschitz, Lukas and Shokri, Reza and Brinton, Christopher G. and Sim, Robert},
  booktitle={The Thirty-Ninth Annual Conference on Neural Information Processing Systems},
  year={2025},
  eprint={2506.04245},
  archivePrefix={arXiv},
  primaryClass={cs.AI}
}

@misc{li2026prefixunderstandadaptuser,
  title={{PrefIx}: Understand and Adapt to User Preference in Human-Agent Interaction},
  author={Li, Jialin and Chen, Zhenhao and Luo, Hanjun and Salam, Hanan},
  year={2026},
  eprint={2602.06714},
  archivePrefix={arXiv},
  primaryClass={cs.HC},
  doi={10.48550/arXiv.2602.06714},
  url={https://doi.org/10.48550/arXiv.2602.06714}
}

@article{jiang2026scribe,
  title={{SCRIBE}: Structured Mid-Level Supervision for Tool-Using Language Models},
  author={Jiang, Yuxuan and Ferraro, Francis},
  journal={arXiv preprint arXiv:2601.03555},
  year={2026},
  eprint={2601.03555},
  archivePrefix={arXiv},
  primaryClass={cs.AI},
  doi={10.48550/arXiv.2601.03555},
  url={https://doi.org/10.48550/arXiv.2601.03555}
}

@article{lin2026reflect,
  title={{Reflect-Guard}: Enhancing {LLM} Safeguards against Adversarial Prompts via Logical Self-Reflection},
  author={Lin, Lixing and You, Juli and Li, Yue and Lin, Luyun and Wang, Yiqing and Zhang, Zhen and Zheng, Moxuan},
  journal={arXiv preprint arXiv:2605.24834},
  year={2026},
  eprint={2605.24834},
  archivePrefix={arXiv},
  primaryClass={cs.CR},
  doi={10.48550/arXiv.2605.24834},
  url={https://doi.org/10.48550/arXiv.2605.24834}
}

@misc{cheng2026resolvingrobustnessprecisiontradeofffinancial,
  title={Resolving the Robustness-Precision Trade-off in Financial {RAG} through Hybrid Document-Routed Retrieval},
  author={Cheng, Zhiyuan and Lai, Longying and Liu, Yue},
  year={2026},
  eprint={2603.26815},
  archivePrefix={arXiv},
  primaryClass={cs.CL},
  doi={10.48550/arXiv.2603.26815},
  url={https://doi.org/10.48550/arXiv.2603.26815}
}

@misc{liu2026improvingcompletenesscomparabilitysegment,
  title={Improving the Completeness and Comparability of Segment Disclosures: A Large Language Model Approach},
  author={Liu, Yue and Cheng, Zhiyuan and Lai, Longying},
  year={2026},
  eprint={2605.23924},
  archivePrefix={arXiv},
  primaryClass={cs.CL},
  doi={10.48550/arXiv.2605.23924},
  url={https://doi.org/10.48550/arXiv.2605.23924}
}

@article{tao2026grasp,
  title={{GRASP}: Plan-Guided Graph Retrieval with Adaptive Fusion and Reranking on Semi-Structured Knowledge Bases},
  author={Tao, Yicheng and Wang, Yiqun and Song, Xiangchen and Luo, Xin and Liu, Kai and Liu, Jie},
  journal={arXiv preprint arXiv:2605.30237},
  year={2026},
  eprint={2605.30237},
  archivePrefix={arXiv},
  primaryClass={cs.IR},
  doi={10.48550/arXiv.2605.30237},
  url={https://doi.org/10.48550/arXiv.2605.30237}
}

@article{lan2026alternating,
  title={Alternating Reinforcement Learning with Contextual Rubric Rewards: Beyond the Scalarization Strategy},
  author={Lan, Guangchen and Xiong, Lian and Zhou, Xin and Cui, Hejie and Zhang, Yuwei and Li, Mao and Shi, Zhenyu and Fetahu, Besnik and Li, Lihong and Li, Xian},
  journal={arXiv preprint arXiv:2603.15646},
  year={2026},
  eprint={2603.15646},
  archivePrefix={arXiv},
  primaryClass={cs.LG},
  doi={10.48550/arXiv.2603.15646},
  url={https://doi.org/10.48550/arXiv.2603.15646}
}

@misc{sun2026beyond,
  title={Beyond Accuracy: Measuring Bias Acknowledgment in Chain-of-Thought Reasoning for Responsible {AI} Evaluation},
  author={Sun, Xian and Gao, Wei and Wang, Yingshuo and Kong, Lingdong and Li, Yanhang and Fan, Zhichao and Zhuang, Zexin and Dong, Wenlong and Zheng, Zhiyuan and Paranjape, Hrishikesh and Mandal, Abhishek and Zhang, Johnny R.},
  year={2026},
  eprint={2606.15127},
  archivePrefix={arXiv},
  primaryClass={cs.LG},
  doi={10.48550/arXiv.2606.15127},
  url={https://doi.org/10.48550/arXiv.2606.15127},
  note={ICML 2026 Workshop on Trustworthy AI for Good}
}

@misc{wang2026timeseriesfoundationmodel,
  title={Do Time Series Foundation Model Benchmarks Hide Regime-Dependent Failures? Evidence from Traffic Speed Forecasting},
  author={Wang, Yingshuo and Sun, Xian and Kong, Lingdong and Gao, Wei and Li, Yanhang and Fan, Zhichao and Zhuang, Zexin},
  year={2026},
  eprint={2606.18367},
  archivePrefix={arXiv},
  primaryClass={cs.LG},
  doi={10.48550/arXiv.2606.18367},
  url={https://doi.org/10.48550/arXiv.2606.18367},
  note={Accepted at the Workshop on Forecasting as a New Frontier of Intelligence, ICML 2026}
}

@misc{wang2026embeddingfoundationmodelpredictions,
  title={Embedding Foundation Model Predictions in Discrete-Choice Models with Structural Guarantees},
  author={Wang, Yingshuo and Sun, Xian and Li, Yanhang and Fan, Zhichao and Zhuang, Zexin},
  year={2026},
  eprint={2606.26432},
  archivePrefix={arXiv},
  primaryClass={cs.LG},
  doi={10.48550/arXiv.2606.26432},
  url={https://doi.org/10.48550/arXiv.2606.26432}
}

@misc{han2026earlyearlyenoughdesigndependent,
  title={How Early Is Early Enough? Design-Dependent Observation-Window Sufficiency in Subscription Churn Prediction},
  author={Han, Xiao and Xiao, Yao and Wu, Chenyu and Zhang, Tongchen},
  year={2026},
  eprint={2607.00473},
  archivePrefix={arXiv},
  primaryClass={cs.LG},
  doi={10.48550/arXiv.2607.00473},
  url={https://doi.org/10.48550/arXiv.2607.00473}
}

@misc{wu2026classweightingversusconditioning,
  title={Class Weighting versus Amount Conditioning in Credit-Card Fraud Detection: A Dollar-Metric Study with a Temporal Explanation Audit},
  author={Wu, Chenyu},
  year={2026},
  eprint={2607.14686},
  archivePrefix={arXiv},
  primaryClass={cs.CE},
  doi={10.48550/arXiv.2607.14686},
  url={https://doi.org/10.48550/arXiv.2607.14686}
}

\end{document}